\documentclass[twoside,leqno,twocolumn]{article}

\usepackage[letterpaper]{geometry}

\usepackage{siamproceedings}

\usepackage[T1]{fontenc}
\usepackage{amsfonts}
\usepackage{graphicx}
\usepackage{epstopdf}
\usepackage{enumitem}
\usepackage{algorithmic}
\usepackage{booktabs}
\usepackage{xcolor}
\usepackage{multirow}
\ifpdf
  \DeclareGraphicsExtensions{.eps,.pdf,.png,.jpg}
\else
  \DeclareGraphicsExtensions{.eps}
\fi

\newsiamremark{remark}{Remark}
\newsiamremark{hypothesis}{Hypothesis}
\crefname{hypothesis}{Hypothesis}{Hypotheses}
\newsiamthm{claim}{Claim}

\usepackage{amsopn}

\begin{document}

\newcommand\relatedversion{}
\renewcommand\relatedversion{\thanks{The full version of the paper can be accessed at \protect\url{https://arxiv.org/abs/0000.00000}}} 

\title{\Large Are LLM Uncertainty and Correctness Encoded by the Same Features? \\A Functional Dissociation via Sparse Autoencoders}
    \author{
  Het Patel\thanks{University of California, Riverside, CA
    (\email{hpate061@ucr.edu}).}
  \and Tiejin Chen\thanks{Arizona State University, Tempe, AZ.}
  \and Hua Wei\footnotemark[2]
  \and Evangelos E. Papalexakis\footnotemark[1]
  \and Jia Chen\footnotemark[1]
}

\date{}

\maketitle

\begin{abstract}
Large language models can be uncertain yet correct, or confident yet wrong, raising the question of whether their output-level uncertainty and their actual correctness are driven by the same internal mechanisms or by distinct feature populations. We introduce a $2\times2$ framework that partitions model predictions along correctness and confidence axes, and uses sparse autoencoders to identify features associated with each dimension independently. Applying this to Llama-3.1-8B and Gemma-2-9B, we identify three feature populations that play fundamentally different functional roles. \textbf{Pure uncertainty features} are functionally essential: suppressing them severely degrades accuracy. \textbf{Pure incorrectness features} are functionally inert: despite showing statistically significant activation differences between correct and incorrect predictions, the majority produce near-zero change in accuracy when suppressed. \textbf{Confounded features} that encode both signals are detrimental to output quality, and targeted suppression of them yields a 1.1\% accuracy improvement and a 75\% entropy reduction, with effects transferring across the ARC-Challenge and RACE benchmarks. The feature categories are also informationally distinct: the activations of just 3 confounded features from a single mid-network layer predict model correctness (AUROC $\sim$0.79), enabling selective abstention that raises accuracy from 62\% to 81\% at 53\% coverage. The results demonstrate that uncertainty and correctness are distinct internal phenomena, with implications for interpretability and targeted inference-time intervention. 

\end{abstract}


\section{Introduction.}

\begin{figure}[h!]
    \centering
    \includegraphics[width=\linewidth]{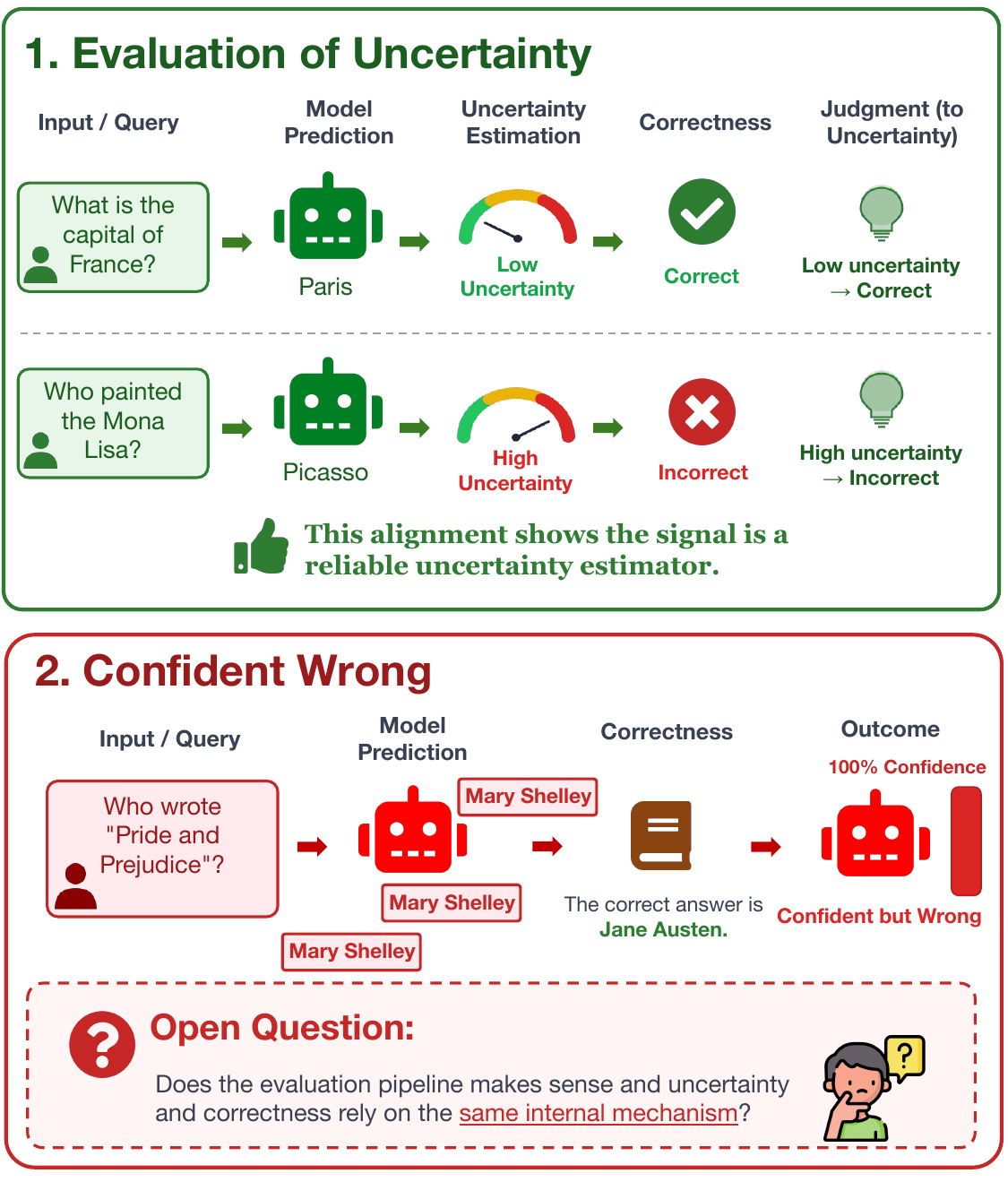}
    \vspace{-6pt}
    \caption{\textbf{Uncertainty as a signal for correctness, and where it fails.} \textit{Top:} uncertainty aligns with reliability: the model is confident when correct, uncertain when incorrect. \textit{Bottom:} a failure case where the model is confidently wrong. Both are common in deployment, raising our open question: when a model is uncertain or wrong, are these signals driven by the same internal mechanism, or by distinct feature populations?}
    
    \label{fig:Framework}
\end{figure}


As large language models (LLMs) are increasingly deployed in high-stakes settings~\cite{cheong2024not, singh2025openai, yang2025qwen3,li2023large}, determining
when a model's output should be trusted has become a central concern. To address it, various uncertainty quantification (UQ) methods have been proposed while being evaluated using a shared criterion ~\cite{kuhn2023semanticuncertaintylinguisticinvariances, lin2024generatingconfidenceuncertaintyquantification, da2024llm, chen2025uncertaintyquantificationlargelanguage}, 
i.e., a good UQ should discriminate between correct and incorrect predictions, typically measured through metrics such as AUROC and AUPRC~\cite{lin2024generatingconfidenceuncertaintyquantification, kuhn2023semanticuncertaintylinguisticinvariances}. The implicit assumption behind this protocol is that uncertainty and correctness are tightly coupled, and that effective uncertainty estimation is correlated
with predicting whether the model will be correct. The cases when a model is uncertain yet correct, or confident yet wrong, have been well documented ~\cite{huang2025survey}, which
suggests that the observed correlation between uncertainty and correctness may not reflect a shared internal mechanism. Instead, the model may encode these two signals (i.e., correctness and uncertainty) through separate feature populations.
If this is the case, the standard UQ evaluation paradigm conflates two fundamentally independent phenomena: a model's confidence in its prediction and the prediction's actual correctness. This conflation carries practical consequences. An intervention designed to reduce uncertainty would not necessarily improve correctness, and a high AUROC during evaluation would not guarantee that the uncertainty signal and the correctness signal originate from the same internal computation.


A growing body of work on internal representations lends plausibility to this concern. Studies have shown that LLMs encode information about their own correctness~\cite{kadavath2022languagemodelsmostlyknow, burns2024discoveringlatentknowledgelanguage, azaria2023internalstatellmknows} and knowledge awareness~\cite{ferrando2025iknowentityknowledge} in their hidden states, and that inference-time interventions on these representations can steer model behavior~\cite{li2024inferencetimeinterventionelicitingtruthful}. Yet these works study either uncertainty or correctness in isolation. None has tested whether the internal features associated with uncertainty and those associated with correctness are the same features, or whether they can be independently identified and controlled, leaving a practical gap in the domain.

In this paper, we address this question by using sparse autoencoders (SAEs) to decompose LLM internal representations into individual features and testing whether uncertainty and correctness are encoded by distinct or overlapping feature populations. We introduce a $2\times2$ quadrant framework that partitions model predictions along two independent axes, correctness and confidence, and applies the same statistical test to both axes to identify features associated with each signal. This procedure yields three distinct feature populations: pure uncertainty features, pure incorrectness features, and confounded features 
associated with both signals. We then conduct targeted suppression experiments to determine the functional role of each population, revealing a three-way dissociation in which each category plays a fundamentally different role in the model's computation. Figure~\ref{fig:Framework} shows the overview of our proposed work. Our contributions are as follows:
\begin{enumerate}
    \item \textbf{A $2\times2$ quadrant framework} that disentangles uncertainty from incorrectness at the level of individual SAE features, enabling independent analysis of each signal.
    
    \item \textbf{A three-way functional dissociation} showing that pure uncertainty features are essential (suppression degrades accuracy), pure incorrectness features are inert (most produce near-zero accuracy change), and confounded features are detrimental to output quality (their suppression improves accuracy by 1.1\% and reduces entropy by 75\%).

    
    \item \textbf{Cross-model and cross-dataset generalization} demonstrating that the functional dissociation replicates across Llama-3.1-8B and Gemma-2-9B, and that features discovered on MMLU  transfer to ARC-Challenge and RACE without reselection.
    \item \textbf{A predictive dissociation} showing that just 3 of  102 confounded features from a single mid-network layer predict model correctness, enabling selective abstention that raises accuracy from 62\% to 81\% at 53\% question coverage. 
\end{enumerate}

\section{Related Work.}

\subsection{Uncertainty quantification in LLMs.} Existing UQ methods primarily operate on model outputs. Kuhn et al. \cite{kuhn2023semanticuncertaintylinguisticinvariances} introduce semantic entropy, which clusters generated responses by meaning to estimate uncertainty. Lin et al. \cite{lin2024generatingconfidenceuncertaintyquantification} use NLI-based similarity and graph Laplacian methods for black-box UQ. Chen et al. \cite{chen2025uncertaintyquantificationlargelanguage} propose a multi-dimensional framework that integrates semantic and knowledge-aware similarity using tensor decomposition. While effective for flagging unreliable outputs, these approaches cannot determine whether the model's expressed uncertainty and its prediction errors originate from shared or distinct internal mechanisms---a question that requires operating on internal representations rather than outputs alone.


\subsection{Internal representations of correctness.} A growing body of work demonstrates that LLMs encode signals about their own knowledge and correctness internally. Kadavath et al. \cite{kadavath2022languagemodelsmostlyknow} show that language models can predict their correctness via self-evaluation prompts that elicit calibrated confidence scores, demonstrating internal calibration. Burns et al. \cite{burns2024discoveringlatentknowledgelanguage} discover latent truth directions in activation space through unsupervised consistency constraints, and Azaria and Mitchell \cite{azaria2023internalstatellmknows} train classifiers on hidden states to distinguish true from false statements. These studies establish that models maintain internal representations relevant to correctness, but they typically operate along a single axis, probing for "truth" or "correctness" without distinguishing between features associated with uncertainty and those associated with prediction errors. 

\subsection{Sparse autoencoders for interpretability.} Sparse autoencoders decompose polysemantic neural activations into interpretable features, grounded in the theory of superposition \cite{elhage2022toymodelssuperposition} and validated through dictionary-learning recoveries \cite{bricken2023monosemanticity, cunningham2023sparseautoencodershighlyinterpretable}. Ferrando et al. \cite{ferrando2025iknowentityknowledge} use SAEs on Gemma-2 to identify features encoding entity recognition, and show that steering them can induce or suppress hallucinations. Our work uses similar tools but addresses a different question: rather than probing entity knowledge, we disentangle uncertainty from incorrectness and show that the resulting feature categories play fundamentally different functional roles.

\section{Background.} 

SAEs decompose dense neural activations into a higher-dimensional sparse feature space where individual dimensions are more likely to correspond to interpretable concepts \cite{bricken2023monosemanticity, cunningham2023sparseautoencodershighlyinterpretable}. An SAE encodes a residual stream activation $\mathbf{x} \in \mathbb{R}^{d_{\text{model}}}$ into a sparse latent $\mathbf{z} \in \mathbb{R}^m$ with $m \gg d_{\text{model}}$, then decodes it:
\begin{equation}
\mathbf{z} = \sigma(W_{\text{enc}}(\mathbf{x} - \mathbf{b}_{\text{pre}}) + \mathbf{b}_{\text{enc}}), \quad \hat{\mathbf{x}} = W_{\text{dec}}\mathbf{z} + \mathbf{b}_{\text{pre}}
\end{equation}
\noindent trained to minimize reconstruction error $\|\mathbf{x} - \hat{\mathbf{x}}\|^2$ subject to a sparsity penalty on $\mathbf{z}$, with $\sigma$ a sparsity-inducing nonlinearity (e.g., ReLU or TopK). The rows of $W_{\text{dec}}$ serve as feature directions, and each entry of $\mathbf{z}$ measures how strongly that feature is active.

We use pretrained Llama Scope \cite{LlamaScope} ($d_{\text{model}}{=}4{,}096$, $m{=}32{,}768$) for Llama-3.1-8B and Gemma Scope \cite{GemmaScope} ($d_{\text{model}}{=}3{,}584$, $m{=}16{,}384$) for Gemma-2-9B, with one SAE per residual stream layer. All feature discovery and suppression operations operate on these sparse representations. The feature populations (pure uncertainty, pure incorrectness, and confounded) that we identify depend on the SAE architecture and training procedure. We use well-validated published SAEs to minimize this concern and discuss alternative SAE configurations as a limitation in Section \ref{sec: Discussion}.

\section{Proposed Work.}

\subsection{Feature Discovery via Quadrant Analysis.}
We use a $2\times2$ quadrant framework to identify sparse autoencoder features in LLM internal representations that encode uncertainty versus incorrectness. We focus on multiple-choice question (MCQ) datasets because correctness is binary and uncertainty can be measured directly from the answer-token distribution. Open-ended settings require hand-crafted uncertainty estimation (e.g., semantic clustering, NLI-based similarity, sampled-response entropy)~\cite{kuhn2023semanticuncertaintylinguisticinvariances,lin2024generatingconfidenceuncertaintyquantification} and are harder to compare across models because token-level probabilities depend on the vocabulary and tokenization. MCQ gives a clean, model-agnostic measurement substrate for the quadrant analysis. Given a set of MCQs, we run inference through the model and record two quantities per question: a correctness label (whether the model's top prediction among the answer tokens matches the ground truth) and an uncertainty score defined as the Shannon entropy of the softmax distribution over the four answer tokens, $H(p) = -\sum_{i=1}^{4} p_i \log p_i$, where $p_i$ is the model's probability assigned to answer choice $i$. We then partition questions along two axes into four groups (Figure \ref{fig:Framework}, 2b): confident-correct (A), confident-incorrect (B), uncertain-correct (C), and uncertain-incorrect (D).


The confidence axis is determined by thresholding the entropy. In our experiments, we use the 25th  and 75th percentiles of the entropy distribution as thresholds, classifying questions below the 25th percentile as confident and those above the 75th percentile as uncertain. Questions in the middle 50\% are excluded to ensure a clean separation between groups. The 25th/75th percentile cutoff was chosen as a tail-vs-tail contrast that yields clearly separated confident and uncertain groups while retaining sufficient sample size in each group for the Mann-Whitney U test. We assess sensitivity to this threshold in Section \ref{sec: feature_dis} and find that confounded features are highly robust (90.2\% retain their classification at the median), while pure incorrectness features show lower retention (only 7.5\% retain their classification), a point we return to when interpreting the inertness finding. 

 \begin{figure}[h]
    \centering
    \includegraphics[width=\linewidth]{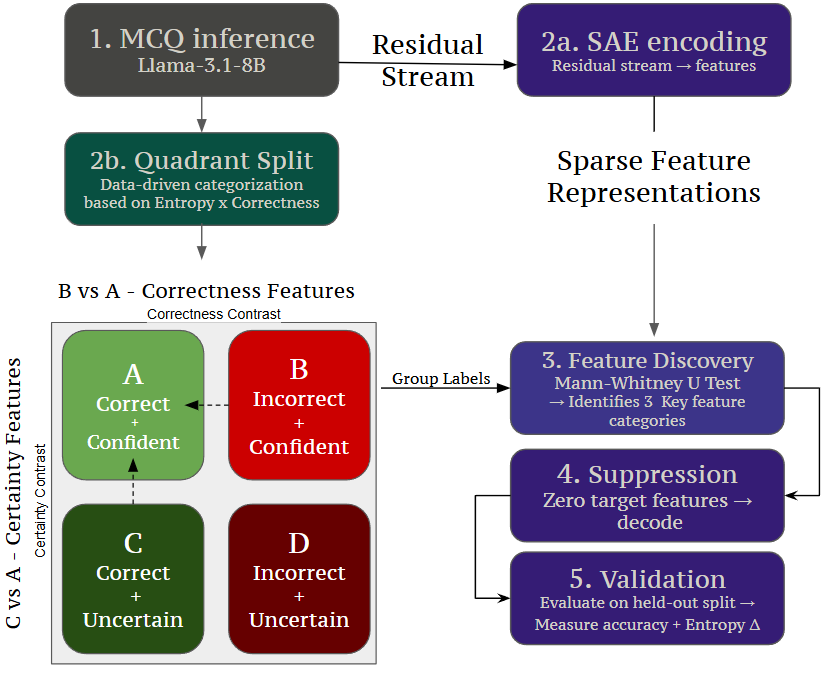}
    \caption{Experimental framework. MCQ inference (1) produces residual stream activations, which are encoded by sparse autoencoders (2a). Predictions and output entropy define a $2\times2$ quadrant split (2b). Feature discovery (3) applies Mann-Whitney U tests to the quadrant groups, yielding three categories. Suppression (4) zeroes selected features; validation (5) evaluates on a held-out split.}
    \label{fig:Framework}
\end{figure}


At each layer, we intercept the residual stream and encode it with a pretrained sparse autoencoder, producing a sparse activation vector in the SAE's latent space. For each feature, we apply the Mann-Whitney U test to two comparisons and compute the effect size (Cohen's $d$) to quantify the between-group difference. For two groups with means $\mu_1, \mu_2$ and pooled standard deviation $s_p$, Cohen's $d = (\mu_1 - \mu_2) / s_p$, measuring how many standard deviations apart the two group means are. We retain only features with $d > 0$, i.e., higher activation in the uncertain or incorrect group. The first comparison, Group C vs. Group A, holds correctness constant while varying confidence. The second, Group B vs. Group A, holds confidence constant while varying correctness. A feature is marked as significant if it reaches $p < 0.05$ with a positive effect size. Features significant only in the first comparison are classified as \textit{pure uncertainty features}. We use "pure" to indicate significance on the uncertainty axis alone among the two we test, not a claim that the features encode uncertainty to the exclusion of all other concepts; such features may still encode other confounds (e.g., question difficulty, domain). Those significant only in the second comparison (i.e., Group B vs. Group A) are \textit{pure incorrectness features}. The features significant in both comparisons are \textit{confounded features}, whose effect size is measured as the minimum of the two comparisons.


\subsection{SAE Suppression Mechanism.}
To test whether the discovered features are functionally relevant to the model's behavior, we intervene on the residual stream using an encode-modify-decode procedure. At the target layer, we encode the residual stream activations at the final-token position through the SAE, zero out the activations of the targeted features, and decode the modified representation. We then compute the difference between the modified and original reconstructions and add this difference to the original residual stream. This ensures that all information not captured by the targeted features is preserved. The model then continues its forward pass with the modified representation. By comparing accuracy and output entropy before and after suppression, we determine whether a feature is essential (suppression hurts accuracy), inert (no effect), or harmful (suppression improves accuracy).

\textbf{SAE reconstruction error control.} As a control, we verify that encoding and decoding through the SAE without suppressing any features yield no change in accuracy or entropy, confirming that the observed effects are attributable to the target features rather than to SAE reconstruction error. 

\subsection{Discovery-Validation Protocol.}
\label{sec: discovery_validation}
To avoid overfitting in feature selection, we use a non-overlapping discovery-validation split. We partition the MMLU \cite{MMLU} test split (14,042 questions) into two equal halves for our discovery and validation subsets ($\sim$7,021 questions each). We use the test split because MMLU's native validation split (1,531 questions) is too small for reliable screening, and we exclude auxiliary\_train to keep cross-dataset transfer (Section \ref{tab:cross_dataset}) clean.



Feature discovery and all screening decisions happen exclusively on the discovery set. For individual suppression testing, we evaluate up to the top 5 features per layer, per category (pure uncertainty, pure incorrectness, or confounded), ranked by effect size. Each feature is suppressed in isolation on the discovery set, and we retain only those for which suppression does not degrade accuracy relative to the unsuppressed model while reducing mean output entropy. We ablate this choice in Section \ref{Ablation} and show that selecting on entropy reduction alone is catastrophic. The features passing screening are then suppressed simultaneously and evaluated on the held-out validation set, which has not been used at any prior stage. Cross-dataset generalization is assessed by applying the same features discovered on MMLU to ARC-Challenge \cite{ARC-Challenge} without re-running the discovery or selection steps. Cross-architecture generalization is evaluated by replicating the full pipeline on Gemma-2-9B \cite{Gemma2} with Gemma Scope SAEs \cite{GemmaScope}.

\section{Results.}

\subsection{Feature Discovery \& Depth Gradient.}
\label{sec: feature_dis}
\label{feature_discovery}
Table \ref{tab:feature_count} summarizes the feature counts identified across all layers on the MMLU discovery set. Across both models, pure uncertainty features are the most numerous category, followed by pure incorrectness, and then confounded. Gemma discovers more features in each category than Llama, likely reflecting differences in SAE training and architecture.

\begin{figure*}[t]
    \centering
    \vspace{-10pt}
    \includegraphics[width=\textwidth]{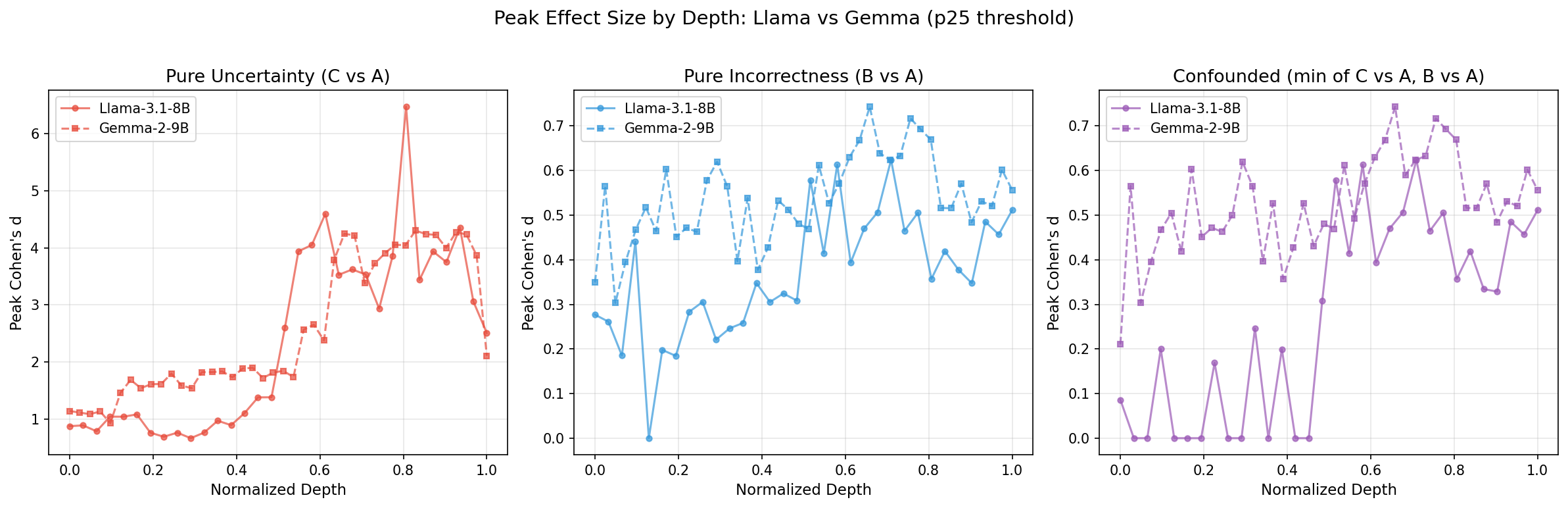}
    \vspace{-10pt}
    \caption{Peak effect size (Cohen's $d$) by normalized depth for three feature categories across Llama-3.1-8B (32 layers) and Gemma-2-9B (42 layers). All three categories show increasing effect sizes with model depth, consistent with greater representational disentangling in later layers. They differ in magnitude: pure uncertainty features (left) reach $d > 6$, while pure incorrectness (center) and confounded features (right) remain below $d = 0.8$. Both models show consistent depth-dependent patterns despite architectural differences. Note: y-axis scales differ across panels.}
    \label{fig:DepthGradient}
    \vspace{-15pt}
\end{figure*}

The three categories differ in the magnitude of effect size. Figure \ref{fig:DepthGradient} plots the effect size $d$ w.r.t.\ normalized depth (layer index divided by total number of layers) for each feature category across both models. Pure uncertainty features reach effect sizes above $d=6$, while pure incorrectness and confounded features remain below $d=0.8$. All three categories show increasing effect sizes with depth: per-feature effects are negligible in the first third of layers (normalized depth $<0.3$), moderate in the middle third, and strongest in the final third. This pattern replicates in both architectures when depth is normalized to [0, 1].



\textbf{Feature categorization robustness to entropy thresholds}. The 25th/75th percentile cutoff isolates the most confidently-separated features on each axis; questions near the median could be assigned to any of the four quadrants and are excluded by design. We assess sensitivity by comparing classifications at this strict cutoff against a looser median split (50th percentile). Under the median split, the pure categories shrink and confounded grows (Llama: 54\% and 65\% drop in pure uncertainty and pure incorrectness, 4.6$\times$ growth in confounded; Gemma: 29\%, 48\%, 2.9$\times$), because features near the boundary cross significance on the second axis once the middle entropy band is included. Of features classified at the 25/75 cutoff, confounded features are most often preserved in their category under the median split (90.2\% Llama, 63.1\% Gemma), followed by pure uncertainty (38.7\%, 61.1\%) and pure incorrectness (7.5\%, 16.4\%). The lower retention for pure incorrectness reflects that its signal is more diffuse than the tail-driven uncertainty and confounded signals: most features migrate into the confounded category rather than dropping out entirely, because the expanded quadrant groups cross significance on the second axis.

Critically, the inertness result in Section \ref{sec:causal_dissociation} is measured on pure-incorrectness features classified under the 25/75 cutoff, which are the most confidently distinguished from the confounded population. These pure-incorrectness features fail to produce behavioral effects under suppression, indicating that inertness is intrinsic to the cleanest pure-incorrectness signal rather than an artifact of sensitivity to thresholds.

\subsection{Functional Dissociation via Suppression.}
\label{sec:causal_dissociation}

\begin{table}[t]
    \centering
    \setlength{\tabcolsep}{4pt} 
    \small 
    \begin{tabular}{l|cc}
        \hline
        & Llama-3.1-8B & Gemma-2-9B \\
        \hline
        No. of uncertainty features   & 700     & 7{,}766 \\
        No. of incorrectness features & 255     & 3{,}778 \\
        No. of confounded features    & 102  & 934     \\
        \hline
    \end{tabular}
    \caption{Feature counts by category on the MMLU discovery set (7K questions, 25th percentile entropy threshold). Gemma discovers more features per category than Llama, likely reflecting differences in SAE training and architecture.}
\label{tab:feature_count}
\end{table}

To test whether the three feature categories correspond to distinct functional roles (e.g., helpful or harmful), we suppress features from each category and measure the effect on model accuracy and output entropy. We test up to 5 features per layer per category (some layers contribute fewer), ranked by effect size, yielding a total number of features tested equal to the sum across layers.

\textbf{Comparison against baselines.} Since feature-level suppression for uncertainty-correctness disentanglement has not been studied previously, we compare against a random-feature control matched in features-per-layer and filtered with the same joint screening criterion, drawn from active features only to avoid dead-feature contamination. Table \ref{tab:main_baseline} shows the effect is specific to the confounded category: on Llama, random suppression stays near baseline (61.96\% / 0.818 vs.\ 61.91\% / 0.824), whereas confounded yields 63.01\% / 0.205; Gemma shows the same pattern (67.63\% / 0.528 random vs.\ 67.85\% / 0.278 confounded). Only 55 features for Llama and 39 for Gemma are suppressed, a tiny fraction of the $\sim$1.05M and $\sim$688K SAE features per model, indicating the effect is driven by a small, functionally specific subset.


\textbf{Pure incorrectness features are functionally inert under suppression}.
Despite showing significantly higher activations on incorrect predictions (the criterion by which they were defined), pure incorrectness features produce no meaningful change in model behavior when suppressed. 
In Llama, of 137 tested features, \textbf{68\%} produced near-zero change in accuracy when suppressed individually, with a mean accuracy delta of \textbf{-0.03\%} and a mean entropy delta of \textbf{+0.001} (essentially zero). Gemma shows the same pattern: of 195 tested features, \textbf{42\%} showed no change in accuracy, with a mean accuracy delta of \textbf{-0.01\%} and a mean entropy delta of \textbf{+0.0003}. These features correlate reliably with incorrectness but produce no measurable behavioral effect under our suppression protocol, though we cannot rule out that these features play functional roles in other tasks or under different interventions.

\textbf{Pure uncertainty features are functionally essential.} Suppressing pure uncertainty features degrades accuracy, with the effect concentrated in late layers. In Llama, the worst single feature causes a $-10.21\%$ accuracy collapse at layer 31, and the surrounding layers consistently produce the largest drops. Gemma shows the same pattern, with a worst-case accuracy drop of $8.70\%$  in the late layers. 


Screening features that individually preserve accuracy (for Llama, 63 of 160 tested candidates pass the screen) and jointly suppressing them yields minimal change in accuracy but a substantial reduction in entropy (-0.38). This is a sanity check on the quadrant framework: even after filtering out destructive features, the remaining pure-uncertainty features still influence output entropy as predicted, confirming that the axis labels track meaningful internal quantities. 


\textbf{Confounded features are detrimental to output quality.} We apply the screening procedure from Section \ref{sec: discovery_validation}. For Llama, 55 of the 102 confounded features pass; for Gemma, the top-5-per-layer cap reduces 934 confounded features to 198 candidates, of which 39 pass. Suppressing the selected features on the held-out validation set yields the headline gains in Table \ref{tab:main_baseline} (Llama: $+1.10\%$ accuracy, $-0.62$ entropy; Gemma: $+0.20\%$ accuracy, $-0.25$ entropy). Confounded features are the only category whose screened subset meaningfully improves accuracy, and as Section \ref{predictive} shows, whose activations best predict model correctness.

\begin{table}[t]
\centering
\caption{Comparison of confounded feature suppression (ours) against the unsuppressed baseline and a random-feature suppression control. The random control suppresses the same number of features per layer as ours, and is filtered using the same joint screening criterion (accuracy preserved and entropy reduced). Held-out MMLU validation set.}
\vspace{3pt}
\label{tab:main_baseline}
\small
\begin{tabular}{llcc}
\toprule
Model & Method & Acc $\uparrow$  & Ent $\downarrow$ \\
\midrule
\multirow{3}{*}{Llama-3.1-8B} 
  & Baseline      & 61.91\% & 0.824 \\
  & Random      & 61.96\% & 0.818 \\
  & Confounded (\textbf{Ours})     & \textbf{63.01\%} & \textbf{0.205} \\
\midrule
\multirow{3}{*}{Gemma-2-9B}  
  & Baseline     & 67.65\% & 0.529 \\
  & Random      & 67.63\% & 0.528 \\
  & Confounded (\textbf{Ours})     & \textbf{67.85\%} & \textbf{0.278} \\
\bottomrule
\end{tabular}
\end{table}

\label{Ablation}
\textbf{Selection criterion ablation.} Because confounded features are significant on both axes by construction, suppressing them tends to reduce entropy, but entropy reduction alone does not indicate whether probability mass shifts towards the correct answer. Table \ref{tab:ablation} compares three criteria on 102 Llama confounded candidates. Accuracy-only and joint selection yield nearly identical results (+1.18\% and + 1.10\%), confirming that the accuracy gate is the operative filter. Entropy-only selection admits features whose suppression concentrates probability mass on incorrect answers, causing a -10.78\% accuracy collapse. We use the joint criterion as a conservative choice.



\begin{table}[t]
\centering
\caption{Selection criterion ablation for confounded features (Llama-3.1-8B), starting from the 102 confounded candidates (top 5 per layer, Section~\ref{sec: discovery_validation}). \#Feats is the number passing each screening criterion; Acc $\Delta$ and Ent $\Delta$ are changes relative to the unsuppressed baseline on the held-out MMLU validation set. The accuracy gate is the operative filter, and the entropy-only selection is catastrophic.}
\label{tab:ablation}
\vspace{3pt}
\begin{tabular}{lccc}
\hline
Criterion & \#Feats & Acc $\Delta$ & Ent $\Delta$ \\
\hline
Accuracy + Entropy & 55 & +1.10\% & $-$0.6187 \\
Accuracy only & 59 & +1.18\% & $-$0.6180 \\
Entropy only & 86 & $-$10.78\% & $-$0.6025 \\
\hline
\end{tabular}
\end{table}

\subsection{Generalization.}
\subsubsection{Cross-Dataset.}
\label{sec: cross_dataset}
To test whether the selected features capture model-general rather than dataset-specific signals, we apply the confounded features discovered and screened on a source dataset to a held-out target, without rerunning discovery or selection. We evaluate three transfer directions: MMLU $\rightarrow$ ARC-Challenge \cite{ARC-Challenge}, and bidirectional transfer between MMLU and RACE \cite{race}.


\begin{table}[h]
\centering
\caption{Cross-dataset generalization. Confounded features discovered and screened in the source are applied to the target without re-selection. Acc $\Delta$ and Ent $\Delta$ are changes relative to the unsuppressed baseline on the target dataset.}
\label{tab:cross_dataset}
\setlength{\tabcolsep}{4pt}
\begin{tabular}{llrrr}
\toprule
Model & Direction & \#F & Acc $\Delta$ $\uparrow$ & Ent $\Delta$ $\downarrow$ \\
\midrule
\multirow{3}{*}{Llama}
 & MMLU $\to$ ARC   & 55 & +0.81\% & $-$0.495 \\
 & MMLU $\to$ RACE  & 55 & +0.70\% & $-$0.455 \\
 & RACE $\to$ MMLU  & 30 & +0.75\% & $-$0.245 \\
\midrule
\multirow{3}{*}{Gemma}
 & MMLU $\to$ ARC   & 39 & +0.23\% & $-$0.102 \\
 & MMLU $\to$ RACE  & 39 & +0.63\% & $-$0.291 \\
 & RACE $\to$ MMLU  & 53 & +0.16\% & $-$0.040 \\
\bottomrule
\end{tabular}
\end{table}

All six transfer cells yield positive accuracy deltas and entropy reductions (Table \ref{tab:cross_dataset}). ARC-Challenge rules out MMLU-specific overfitting within the factual MCQ format, and MMLU $\leftrightarrow$ RACE extends the test to a larger task-format gap (short-form factual QA vs.\ long-context reading comprehension). Llama transfers strongly in all three directions. Gemma transfers strongly from MMLU but weakly from RACE (+0.16\% vs.\ +0.63\% on the reverse), reflecting the features available for transfer: Gemma's RACE-discovered features have mean effect size 0.24 vs.\ 0.44 for MMLU-discovered features, possibly reflecting SAE dependence (Section \ref{sec: Discussion}), though we cannot isolate the cause here.

\subsubsection{Cross-Architecture.}

The complete pipeline was independently replicated in Gemma-2-9B with Gemma Scope SAEs. Despite differences in model family, layer count (42 vs. 32), SAE dimensionality (16K vs. 32K), and baseline accuracy, all core findings replicate: pure incorrectness features are inert, pure uncertainty features are essential, and confounded features yield accuracy improvements and a major entropy reduction when screened and suppressed. Effect sizes also increase monotonically with normalized layer depth (Figure \ref{fig:DepthGradient}). The three-way dissociation replicates across two architectures of comparable scale, suggesting it may reflect a general organizational principle, though confirmation at substantially different model sizes remains an open question for future work. 


\subsection{Predictive Dissociation.}
\label{predictive}

\begin{figure}[t]
    \centering
    \includegraphics[width=0.9\linewidth]{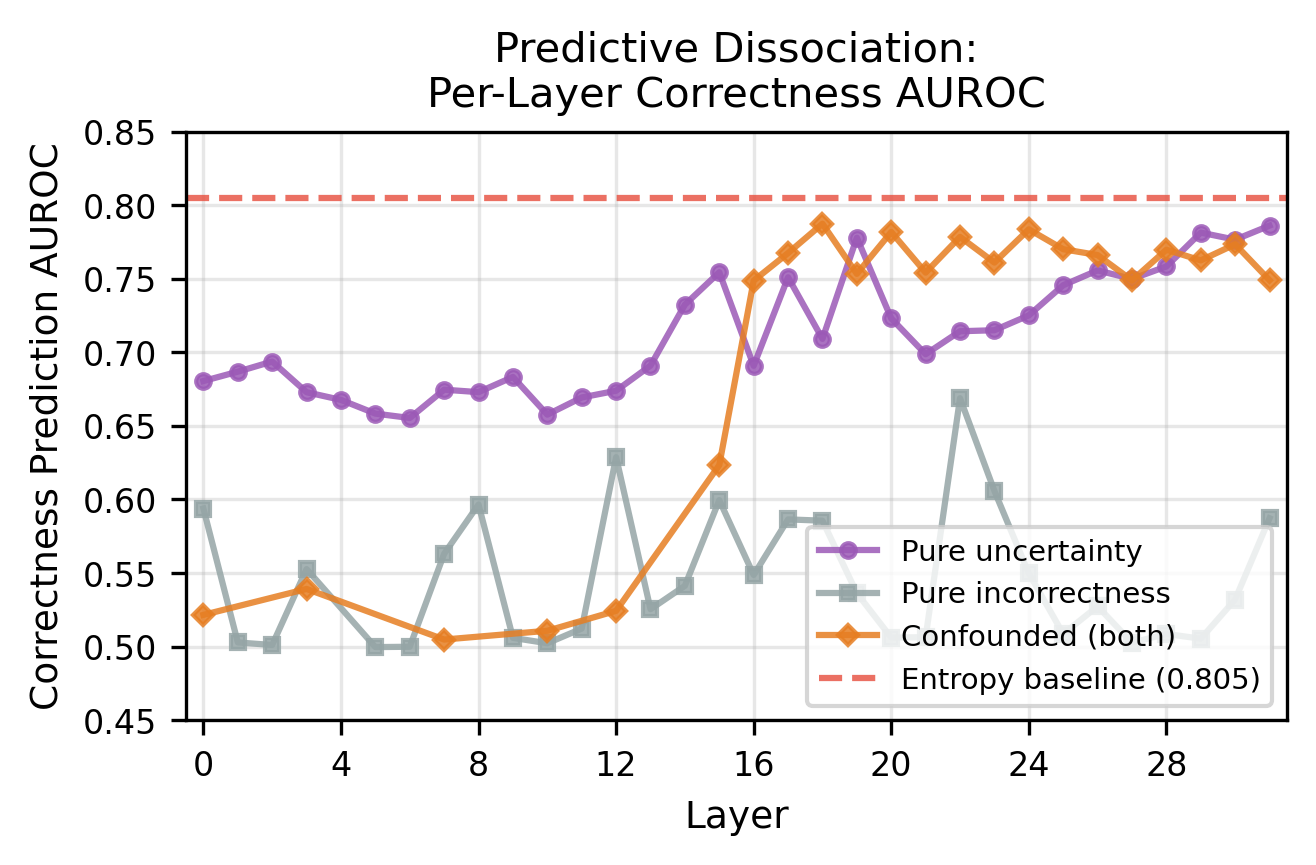}
    \vspace{-10pt}
    \caption{Per-layer correctness prediction AUROC by feature category (Llama-3.1-8B). Per-layer logistic regression classifiers are trained on MMLU discovery set SAE feature activations and evaluated on the held-out validation set. The dashed line shows the AUROC obtained by using output entropy as a predictor of correctness (0.805), which requires a full forward pass.}
    \label{fig:Correctness_prediction}
\end{figure}



\begin{table*}[t]
\centering
\vspace{-10pt}
\caption{Selective abstention using 3 confounded features from layer 18 (highest per-layer AUROC). Classifier trained on MMLU discovery set, evaluated on the held-out validation set. Self-Abstained refers to adding a fifth option, 'I don't know,' and letting the model abstain naturally. Gain is an improvement in accuracy over the 61.91\% baseline (all questions answered).}
\label{tab:abstention}
\begin{tabular}{cccccc}
\toprule
$P(\text{correct}) \geq$ & Answered & Abstained & Coverage & Accuracy & Gain \\
\midrule
0.30 & 6,293 & 728 & 89.6\% & 65.98\% & +4.1\% \\
0.40 & 5,389 & 1,632 & 76.8\% & 70.96\% & +9.1\% \\
0.50 & 4,522 & 2,499 & 64.4\% & 75.90\% & +14.0\% \\
0.60 & 3,709 & 3,312 & 52.8\% & 81.29\% & +19.4\% \\
0.70 & 3,003 & 4,018 & 42.8\% & 86.55\% & +24.6\% \\
\midrule
Self-Abstained  & 6,402 & 619 & 91.2\% & 63.67\% & +1.8\% \\
\bottomrule
\end{tabular}
\end{table*}

The three feature categories are not only functionally distinct (Section~\ref{sec:causal_dissociation}) but also informationally distinct: a classifier trained on each category's activations predicts the binary correctness label with different strengths. We train per-layer logistic regression classifiers on SAE feature activations from the MMLU discovery set (Figure~\ref{fig:Correctness_prediction}). Confounded features emerge as the strongest single-category predictor at layer 16 (normalized depth 0.5) with an AUROC of 0.749, and by layer 18, activation of just 3 confounded features achieves an AUROC of 0.787, approaching the output-entropy baseline (0.805). A classifier trained on MMLU confounded features also transfers to ARC-Challenge without retraining. Pure incorrectness features remain near chance throughout despite showing statistically significant activation differences (Mann-Whitney $p<0.05$), consistent with their inert status under screening and indicating that these features carry faint correlational traces rather than structured representations.

\section{Discussion.}
\label{sec: Discussion}
The three-way dissociation reveals that transformer language models organize uncertainty and incorrectness through functionally distinct feature populations. The same statistical procedure identifies pure uncertainty features and pure incorrectness features with identical thresholds, yet they occupy opposite functional roles. Suppressing uncertainty features destroys accuracy, and suppressing incorrectness features does nothing; only confounded features admit a screened subset whose suppression yields an actionable intervention. The model maintains internal features that reliably track prediction errors, yet suppressing these features produces no measurable change in accuracy or entropy under our intervention protocol.




Confounded features, being the most actionable, may appear to undercut the dissociation claim: if features encoding both signals are the most useful, why insist that uncertainty and incorrectness are distinct? They are identifiable as confounded only because pure uncertainty and pure incorrectness exist as distinct populations under the same procedure. Prior single-axis probes would treat a confounded feature and a pure incorrectness feature as both 'correctness-related'; the three-way partition makes them identifiable and enables targeted interventions that a single-axis approach cannot support. 


\textbf{Depth gradient and superposition.} In early layers, our analysis isolates fewer features with smaller effect sizes, suggesting that individual SAE features there do not cleanly capture uncertainty or correctness signals. This could reflect stronger superposition \cite{Superposition} in those layers, or more generally, more entangled representations. This mirrors weight-level findings: Zhang et al. \cite{LayerImportance} show that removing early cornerstone layers is catastrophic, while LASER \cite{sharma2023truth} and TRAWL \cite{TRAWL} find that rank reduction is most effective in late layers. Early layers resist such decomposition under our method: removing an entire layer is catastrophic, but the individual features our procedure isolates are few and weak, and suppressing them produces negligible behavioral effects.

\textbf{Practical implications.} Confounded features represent the most actionable category for intervention design. Their suppression improves accuracy and substantially reduces output entropy, and these gains transfer across datasets without re-selection. The selective abstention results (Table \ref{tab:abstention}) further demonstrate their utility: 3 features from a single layer enable accuracy-coverage tradeoffs that substantially outperform the model's own self-abstention. This suggests that confounded features encode a compact, transferable signal about prediction reliability that the model does not fully exploit in its output.  

\textbf{Limitations.} Our framework uses a discrete entropy threshold to define the confidence axis (Section \ref{feature_discovery} shows confounded features are robust to this choice). Our evaluation is limited to multiple-choice tasks, and the functional roles may be task-dependent. Features inert under MCQ may participate in chain-of-thought or open-ended generation, though the quadrant framework itself is task-agnostic. Both models are in the 8-9B range; scale generalization is untested. Our conclusions depend on the specific SAE decompositions used (Llama and Gemma Scope). Alternative architectures, dictionary sizes, or sparsity penalties could yield different feature populations. We use published, well-validated SAEs to minimize this concern and leave systematic comparison of SAE variants to future work. The suppression mechanism (encode-modify-decode with zeroing) is one of several possible interventions, and activation patching, steering vectors, or gradient-based methods may reveal relationships that zeroing does not capture. Finally, we test thousands of features with $p<0.05$ without formal multiple-comparison corrections, which could affect the counts in Table \ref{tab:feature_count}.

\section{Conclusion.}
We introduced a $2\times2$ quadrant framework that disentangles uncertainty from incorrectness at the level of individual SAE features and reveals a three-way dissociation. Pure uncertainty features are functionally essential. Pure incorrectness features are inert under suppression, whereas confounded features degrade model output. The model maintains internal features that reliably track prediction errors, yet suppressing these features does not affect the output. Confounded features, encoding both signals, are the most actionable: their suppression reduces entropy by 75\% with a modest accuracy gain, and just 3 such features from a single layer predict correctness within 0.02 AUROC of the entropy baseline. These findings replicate across two architectures of comparable scale, suggesting that the dissociation may reflect a general organizational principle of how transformers encode uncertainty and correctness. Our results suggest that the standard single-axis approach to UQ evaluation may systematically conflate functionally distinct internal phenomena, and that the mechanistic interpretability toolkit can be leveraged to disentangle them.

Future directions include scaling to larger models (70B+) to test whether the three-way dissociation persists and extending the analysis to open-ended generation, where features inert under MCQ may participate in computation that the multiple-choice format does not exercise. The compact predictive signal in confounded features also suggests internal-feature-based abstention as a foundation for more reliable inference-time decision making, especially when combined with existing UQ methods.

\section*{Acknowledgments.}
The authors acknowledge the use of AI-based writing assistance during manuscript preparation. Grammarly and Claude were used to check grammar and improve the clarity and conciseness of the authors' draft.

\newpage
\appendix


\bibliographystyle{siamplain}
\bibliography{bib}
\end{document}